\documentclass{article}

\usepackage{microtype}
\usepackage{graphicx}
\usepackage{subfigure}
\usepackage{booktabs} % for professional tables

\usepackage{amsmath}
\usepackage{amsfonts}

\usepackage{float}
\usepackage{tabularx}
\usepackage{soul}

\newcommand{\kmat}{\mathbf{K}}

\newcommand{\pmat}{\mathbf{P}}
\newcommand{\qmat}{\mathbf{Q}}
\newcommand{\rmat}{\mathbf{R}}

\newcommand{\wmat}{\mathbf{W}}

\newcommand{\eye}{\mathbf{I}}

\newcommand{\gvec}{\mathbf{g}}

\newcommand{\kvec}{\mathbf{k}}
\newcommand{\mvec}{\mathbf{m}}
\newcommand{\svec}{\mathbf{s}}

\newcommand{\vvec}{\mathbf{v}}
\newcommand{\wvec}{\mathbf{w}}
\newcommand{\xvec}{\mathbf{x}}

\newcommand{\zvec}{\mathbf{z}}

\newcommand{\betavec}{\boldsymbol{\beta}}
\newcommand{\gammavec}{\boldsymbol{\gamma}}

\newcommand{\lbar}{\left|}
\newcommand{\rbar}{\right|}

\usepackage{hyperref}

\usepackage[accepted]{icml2019}

\makeatletter
\newcommand\footnoteref[1]{\protected@xdef\@thefnmark{\ref{#1}}\@footnotemark}
\makeatother

\icmltitlerunning{Emerging Convolutions for Generative Normalizing Flows}

\begin{document}

\twocolumn[
\icmltitle{Emerging Convolutions for Generative Normalizing Flows}




\begin{icmlauthorlist}
\icmlauthor{Emiel Hoogeboom}{bosch}
\icmlauthor{Rianne van den Berg}{uva}
\icmlauthor{Max Welling}{bosch,cifar}
% \icmlauthor{Cieua Vvvvv}{goo}
% \icmlauthor{Iaesut Saoeu}{ed}
% \icmlauthor{Fiuea Rrrr}{to}
% \icmlauthor{Tateu H.~Yasehe}{ed,to,goo}
% \icmlauthor{Aaoeu Iasoh}{goo}
% \icmlauthor{Buiui Eueu}{ed}
% \icmlauthor{Aeuia Zzzz}{ed}
% \icmlauthor{Bieea C.~Yyyy}{to,goo}
% \icmlauthor{Teoau Xxxx}{ed}
% \icmlauthor{Eee Pppp}{ed}
\end{icmlauthorlist}

\icmlaffiliation{bosch}{UvA-Bosch Delta Lab, University of Amsterdam, Netherlands}
\icmlaffiliation{uva}{University of Amsterdam, Netherlands}
\icmlaffiliation{cifar}{Canadian Institute for Advanced Research (CIFAR)}
% \icmlaffiliation{goo}{Googol ShallowMind, New London, Michigan, USA}
% \icmlaffiliation{ed}{School of Computation, University of Edenborrow, Edenborrow, United Kingdom}

\icmlcorrespondingauthor{Emiel Hoogeboom}{e.hoogeboom@uva.nl}
% \icmlcorrespondingauthor{Eee Pppp}{ep@eden.co.uk}

\icmlkeywords{invertible convolutions, generative flows, normalizing flows, deconvolution}

\vskip 0.3in
]



\printAffiliationsAndNotice{}  % leave blank if no need to mention equal contribution

\begin{abstract}
Generative flows are attractive because they admit exact likelihood optimization and efficient image synthesis. Recently, \citet{kingma2018glow} demonstrated with Glow that generative flows are capable of generating high quality images. We generalize the $1 \times 1$ convolutions proposed in Glow to \textit{invertible} $d \times d$ convolutions, which are more flexible since they operate on both channel and spatial axes. We propose two methods to produce invertible convolutions that have receptive fields identical to standard convolutions: \textit{Emerging} convolutions are obtained by chaining specific autoregressive convolutions, and \textit{periodic} convolutions are decoupled in the frequency domain. Our experiments show that the flexibility of $d \times d$ convolutions significantly improves the performance of generative flow models on galaxy images, CIFAR10 and ImageNet. 
\end{abstract}

\author{Emiel Hoogeboom, Rianne van den Berg, Max Welling}

\section{Introduction}
%!TEX root = ../main.tex

Generative models aim to learn a representation of the data $p(x)$, in contrast with discriminative models that learn a probability distribution of labels given data $p(y|x)$. Generative modeling may be used for numerous applications such as anomaly detection, denoising, inpainting, and super-resolution. The task of generative modeling is challenging, because data is often very high-dimensional, which makes optimization and choosing a successful objective difficult.

\begin{figure}
    \centering
    \includegraphics[width=0.49\textwidth]{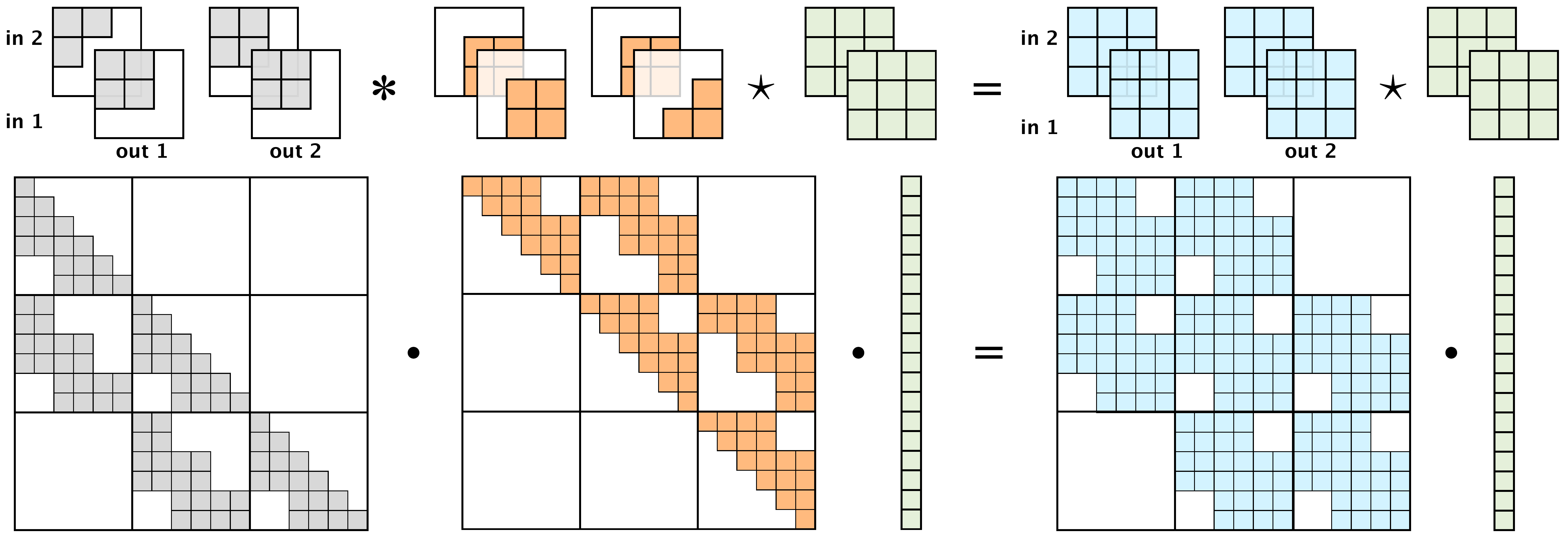}
    \caption{Illustration of a square emerging convolution. The input has spatial dimensions equal to $3 \times 3$, and two channels, white squares denote zero values. Convolutions use one-pixel-wide zero padding at each border. Two consecutive square autoregressive convolutions with filters $k_2$ and $k_1$ have a receptive field identical to a standard convolution, with filter $k_2 *_l k_1$, where $*_l$ denotes a convolution layer. These operations are equivalent to the multiplication of matrices $\text{K}_2 \, \cdot \, \text{K}_1$ and a vectorized input signal $\vec{x}$. Since the filters are learned decomposed, the Jacobian determinant and inverse are straightforward to compute.}
    \label{fig:overview}
    \vspace{-0.5cm}
\end{figure}

\noindent Generative models based on normalizing flows \cite{rippel2013high} have several advantages over other generative models: \textit{i)} They optimize the log likelihood of a continuous distribution exactly, as opposed to Variational Auto-Encoders (VAEs) \citep{kingma2014stochastic, rezende2015stochbackprop} which optimize a lower bound to the log-likelihood.  \textit{ii)} Drawing samples has a computational cost comparable to inference, in contrast with PixelCNNs \citep{van2016pixel}. \textit{iii)} Generative flows also have the potential for huge memory savings, because activations necessary in the backward pass can be obtained by computing the inverse of layers \cite{gomez2017reversible,litraining}.

\noindent The performance of flow-based generative models can be largely attributed to Masked Autoregressive Flows (MAFs) \citep{papamakarios2017masked} and the coupling layers introduced in NICE and RealNVP \citep{dinh2014nice, dinh2016density}. MAFs contain flexible autoregressive transformations, but are computationally expensive to invert, which is a disadvantage for sampling high-dimensional data. Coupling layers transform a subset of the dimensions of the data, parameterized by the remaining dimensions. The inverse of coupling layers is straightforward to compute, which makes them suitable for generative flows. However, since coupling layers can only operate on a subset of the dimensions of the data, they may be limited in flexibility.

To improve their effectiveness, coupling layers are alternated with less complex transformations that do operate on all dimensions of the data. \citet{dinh2016density} use a fixed channel permutation in Real NVP, and \citet{kingma2018glow} utilize learnable $1 \times 1$ convolutions in Glow.

However, $1 \times 1$ convolutions suffer from limited flexibility, and using standard convolutions is not straightforward as they are very computationally expensive to invert. We propose two methods to obtain easily invertible and flexible convolutions: \textit{emerging} and \textit{periodic} convolutions. Both of these convolutions have receptive fields identical to standard convolutions, resulting in flexible transformations over both the channel \textit{and} spatial axes.

The structure of an emerging convolution is depicted in Figure \ref{fig:overview}, where the top depicts the convolution filters, and the bottom shows the equivalent matrices of these convolutions. Two autoregressive convolutions are chained to obtain an emerging receptive field identical to a standard convolution. Empirically, we find that replacing $1 \times 1$ convolutions with the generalized invertible convolutions produces significantly better results on galaxy images, CIFAR10 and ImageNet, even when correcting for the increase in parameters.

In addition to invertible convolutions, we also propose a QR decomposition for $1 \times 1$ convolutions, which resolves flexibility issues of the PLU decomposition proposed by \citet{kingma2018glow}. 

The main contributions of this paper are: \textit{1)} Invertible emerging convolutions using autoregressive convolutions. \textit{2)} Invertible periodic convolutions using decoupling in the frequency domain. \textit{3)} Numerically stable and flexible $1 \times 1$ convolutions parameterized by a QR decomposition. \textit{4)} An accelerated inversion module for autoregressive convolutions. The code is available at: {\small \url{github.com/ehoogeboom/emerging}}.

%\begin{enumerate}
%    \item Invertible emerging convolutions using autoregressive convolutions.
%    \item Invertible periodic convolutions using decoupling in the frequency domain.
%    \item Numerically stable and flexible $1 \times 1$ convolutions parameterized by a QR decomposition.
%    \item An accelerated inversion module for autoregressive convolutions.
%\end{enumerate}

%This paper is structured as follows: Section \ref{sec:background} provides background information related to generative flows. Section \ref{sec:method} presents our methods and in section \ref{sec:related_work} related work is discussed. Section \ref{sec:results} describes our results.

\section{Background}
\label{sec:background}
%!TEX root = ../main.tex

\begin{table*}[t]
\scriptsize
\caption{The definition of several generative normalizing flows. All flow functions have an inverse and determinant that are straightforward to compute. The height $h$, width $w$ and number of channels $n_c$ of an output remains identical to the dimensions of the input. The symbols $\odot$ and $/$ denote element-wise multiplication and division. Input and output may be denoted as tensors $\xvec$ and $\zvec$ with dimensions $n_c \times h \times w$. The inputs and outputs may be denoted as one-dimensional vectors $\vec{\xvec}$ and $\vec{\zvec}$ with dimension $n_c \cdot h \cdot w$. Input and output in frequency domain are denoted with $\hat{\xvec}$ and $\hat{\zvec}$, with dimensions $n_c \times h \times w$, where the last two components denote frequencies.}
\label{tab:overview_flows}
\begin{tabularx}{0.999\textwidth}{p{0.15\textwidth} p{0.22\textwidth} p{0.27\textwidth}p{0.31\textwidth}}
\hline
Generative Flow & Function &  Inverse & Log Determinant \\ \hline \hline
Actnorm &
$\vec{\zvec} = \vec{\xvec} \odot \vec{\gammavec} + \vec{\betavec}$ &
$\vec{\xvec} = (\vec{\zvec} - \vec{\betavec}) / \vec{\gammavec}$ & 
$\text{sum}(\log |\vec{\gammavec}|)$
\\ \hline 
Affine coupling & %
$[\vec{\xvec}_a, \vec{\xvec}_b] = \vec{\xvec}$ \newline
$\vec{\zvec}_a = \vec{\xvec}_a \odot f(\vec{\xvec}_b) + g(\vec{\xvec}_b)$ \newline
$\vec{\zvec} = [\vec{\zvec}_a, \vec{\xvec}_b]$ &
$[\vec{\zvec}_a, \vec{\zvec}_b] = \vec{\zvec}$ \newline
$\vec{\zvec}_a = (\vec{\zvec}_a - g(\vec{\zvec}_b)) / f(\vec{\zvec}_b) $ \newline
$\vec{\xvec} = [\vec{\zvec}_a, \vec{\xvec}_b]$ &
$\text{sum}(\log |f(\vec{\xvec}_b)|)$
\\ \hline
$1 \times 1$ Conv  &
$\forall ij : \zvec_{:,ij} = \wmat \xvec_{:,ij}$ &
$\forall ij : \xvec_{:,ij} = \wmat^{-1} \zvec_{:,ij}$ &
$h \cdot w \cdot \log | \text{det } \wmat|$ 
\\ \hline 
Emerging Conv &
$\kvec = \wvec_1 \odot \mvec_1$ \newline
$\gvec = \wvec_2 \odot \mvec_2$ \newline
$\zvec = \kvec \star_l (\gvec \star_l \xvec)$ &
$\forall t : \vec{y}_t = (\vec{z}_t - \sum_{i=t+1} G_{t, i} \, \vec{y}_i) / G_{t,t}$ \newline 
$\forall t : \vec{x}_t = (\vec{y}_t - \sum_{i=1}^{t-1} K_{t, i} \, \vec{x}_i) / K_{t,t}$ &
% $\forall t : \vec{x}_t = \frac{\vec{z}_t - \sum_{i=1}^{t-1} K_{t, i} \vec{x}_i}{K_{t,t}}$ &
$ \sum_c \log | {\kvec}_{c,c,m_y,m_x} \, \gvec_{c,c,m_y,m_x} |$
% $ \text{sum}(\log | \text{diag}(\kmat) | )$
%$w \cdot h \, \sum_c \log | k_{c,c,m_y,m_x} |$
\\ \hline
Periodic Conv &
$\forall uv : \hat{\zvec}_{:,uv} = \hat{\wmat}_{uv} \hat{\xvec}_{:,uv}$ &
$\forall uv : \hat{\xvec}_{:,uv} = \hat{\wmat}_{uv}^{-1} \hat{\zvec}_{:,uv}$ &
$\sum_{u,v} \log | \text{det }\hat{\wmat}_{uv}|$ 
\\ \hline 
\end{tabularx}
\end{table*}

\subsection{Change of variables formula}
Consider a bijective map between variables $x$ and $z$. The likelihood of the variable $x$ can be written as the likelihood of the transformation $z = f(x)$ evaluated by $p_Z$, using the change of variables formula:
\begin{equation}
p_X(x) = p_Z(z) \lbar \frac{\partial z}{\partial x} \rbar \ \ ; \ \ z = f(x).
\label{eq:change_of_variables}  
\end{equation}
%The complicated probability density $p_X(x)$ is equal to the probability density $p_Z(z)$ multiplied by the Jacobian determinant, where $p_Z$ is chosen to be tractable. The function $f$ can be learned, but the choice of $f$ is constrained by two practical issues: Firstly, the Jacobian determinant should be tractable. Secondly, to draw samples from $p_X$, the inverse of $f$ should be tractable.
The complicated probability density $p_X(x)$ is equal to the probability density $p_Z(z)$ multiplied by the Jacobian determinant, where $p_Z$ is chosen to be tractable. The function $f$ can be learned, but the choice of $f$ is constrained by two practical issues: Firstly, the Jacobian determinant should be tractable. Secondly, to draw samples from $p_X$, the inverse of $f$ should be tractable. 

\subsubsection{Composition of functions}
A sequence composed of several applications of the change of variables formula is often referred to as a normalizing flow \cite{deco1995decorr, tabak2010density, tabak2013family, rezende2015norm}. 
%Likelihoods are generally optimized in log-space for numerical precision, and deep learning models are structured in layers. 
Let $\{h_l\}_{l=1}^{L}$ be the intermediate representations produced by the layers of a neural network, where $z = h_L$ and $h_0 = x$. The log-likelihood of $x$ is written as the log-likelihood of $z$, and the summation of the log Jacobian determinant of each layer:
\begin{equation}
\log p_X(x) = \log p_Z(z) + \sum_{l=1}^L \log \lbar \frac{\partial h_l}{\partial {h_{l-1}}} \rbar.
\end{equation}
\subsubsection{Dequantization}
We will evaluate our methods with experiments on image datasets, where pixels are discrete-valued from 0 to 255. Since generative flows are continuous density models, they may trivially place infinite mass on discretized bin locations.  Therefore, we use the definition of \citet{theis2016note} that defines the relation between a discrete model $\hat{p}(\hat{x})$ and continuous model $p(x)$ as an integration over bins: ${\hat{p}(\hat{x}) \equiv \int_{[0, 1)^d} p(\hat{x} + u) du}$, where $x = \hat{x} + u$. They further derive a lowerbound to optimize this model with Jensen's inequality, resulting in additive uniform noise for the integer valued pixels from the data distribution $\mathcal{D}$:
\begin{align}
%\begin{split}
    \mathbb{E}_{\hat{x} \sim \mathcal{D}} \left[ \vphantom{\sum} \log \hat{p}(\hat{x}) \right]
    &= \mathbb{E}_{\hat{x} \sim \mathcal{D}} \left[ \log \int_{[0, 1)^d} p(\hat{x}+u) du \right] \\
    &\geq \mathbb{E}_{\hat{x} \sim \mathcal{D}, u \sim \mathcal{U}[0, 1)^d} \left[ \vphantom{\sum} \log p(\hat{x}+u) \right]. \notag
%\end{split}
\end{align}
\subsection{Generative flows}
Generative flows are bijective functions, often structured as deep learning layers, that are designed to have tractable Jacobian determinants and inverses. An overview of several generative flows is provided in Table \ref{tab:overview_flows}, and a description is given below: 

\textbf{Coupling layers} \cite{dinh2016density} split the input in two parts. The output is a combination of a copy of the first half, and a transformation of the second half, parametrized by the first part. As a result, the inverse and Jacobian determinant are straightforward to compute.

\textbf{Actnorm layers} \cite{kingma2018glow} are data dependent initialized layers with scale and translation parameters. They are initialized such that the distribution of activations has mean zero and standard deviation one. Actnorm layers improve training stability and performance.

\textbf{$1 \times 1$ Convolutions} \cite{kingma2018glow} are easy to invert, and can be seen as a generalization of the permutation operations that were used by \citet{dinh2016density}. 1 $\times$ 1 convolutions improve the effectiveness of the coupling layers.

\subsection{Convolutions}
%!TEX root = ../main.tex

\begin{figure}
    \centering
    \includegraphics[width=.40\textwidth]{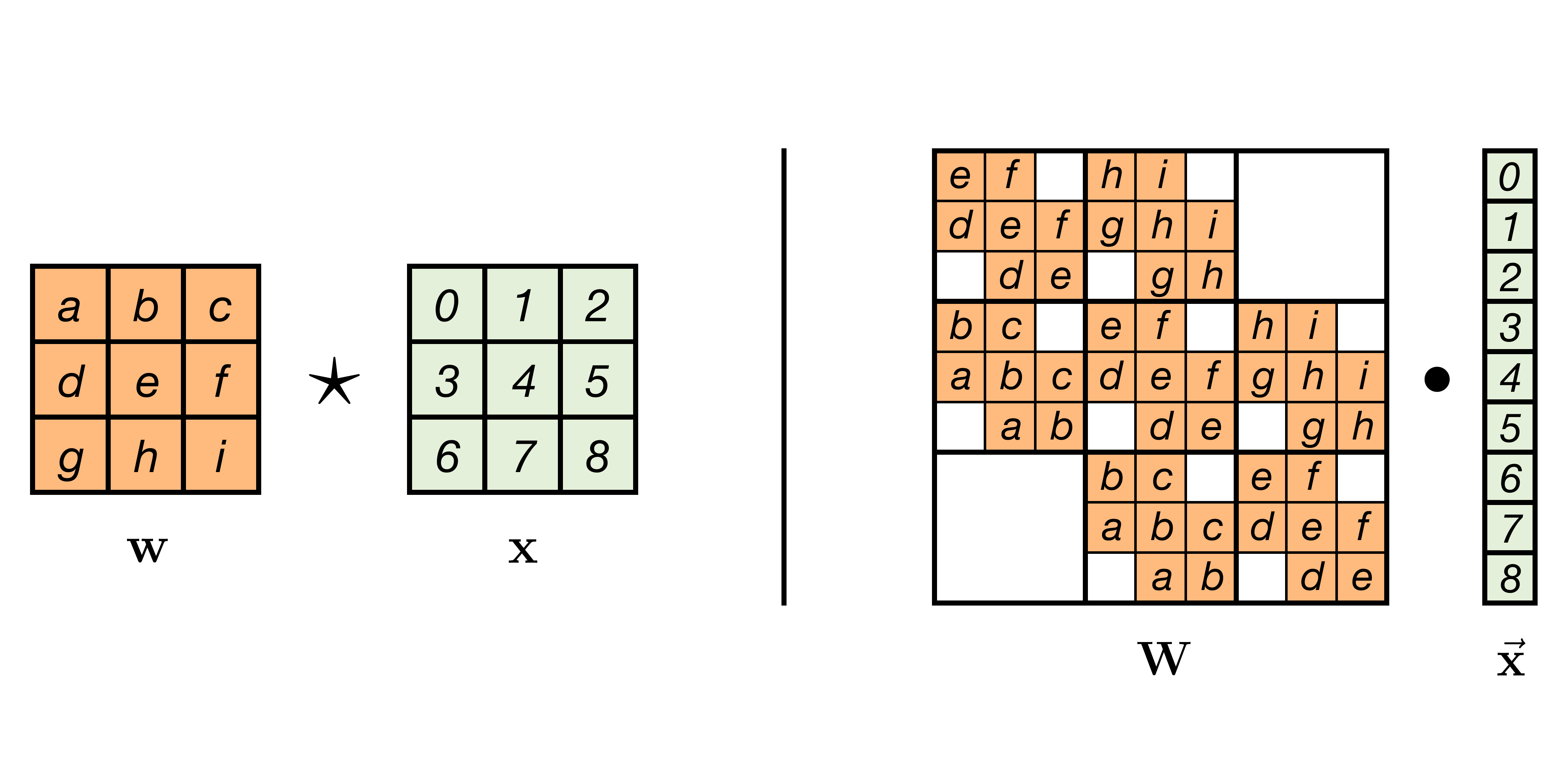}
    % \vspace{-1.26cm}
    \caption{Illustration of a standard 3 $\times$ 3 convolution layer with one input and output channel. The spatial input size is $3 \times 3$, and the input values are $\{0, 1, \dots, 8 \}$. The convolution uses one-pixel-wide zero padding at each border, and the filter has parameters $\{a, b, \dots, i\}$. Left: the convolution $\wvec \star \xvec$. Right: the matrix multiplication $\wmat \cdot \vec{\xvec}$ which produces the equivalent result.}
    \label{fig:convolution_equivalent_matrix_1channel}
    % \vspace{-0.4cm}
\end{figure}

\begin{figure}
    \centering
    \includegraphics[width=.40\textwidth]{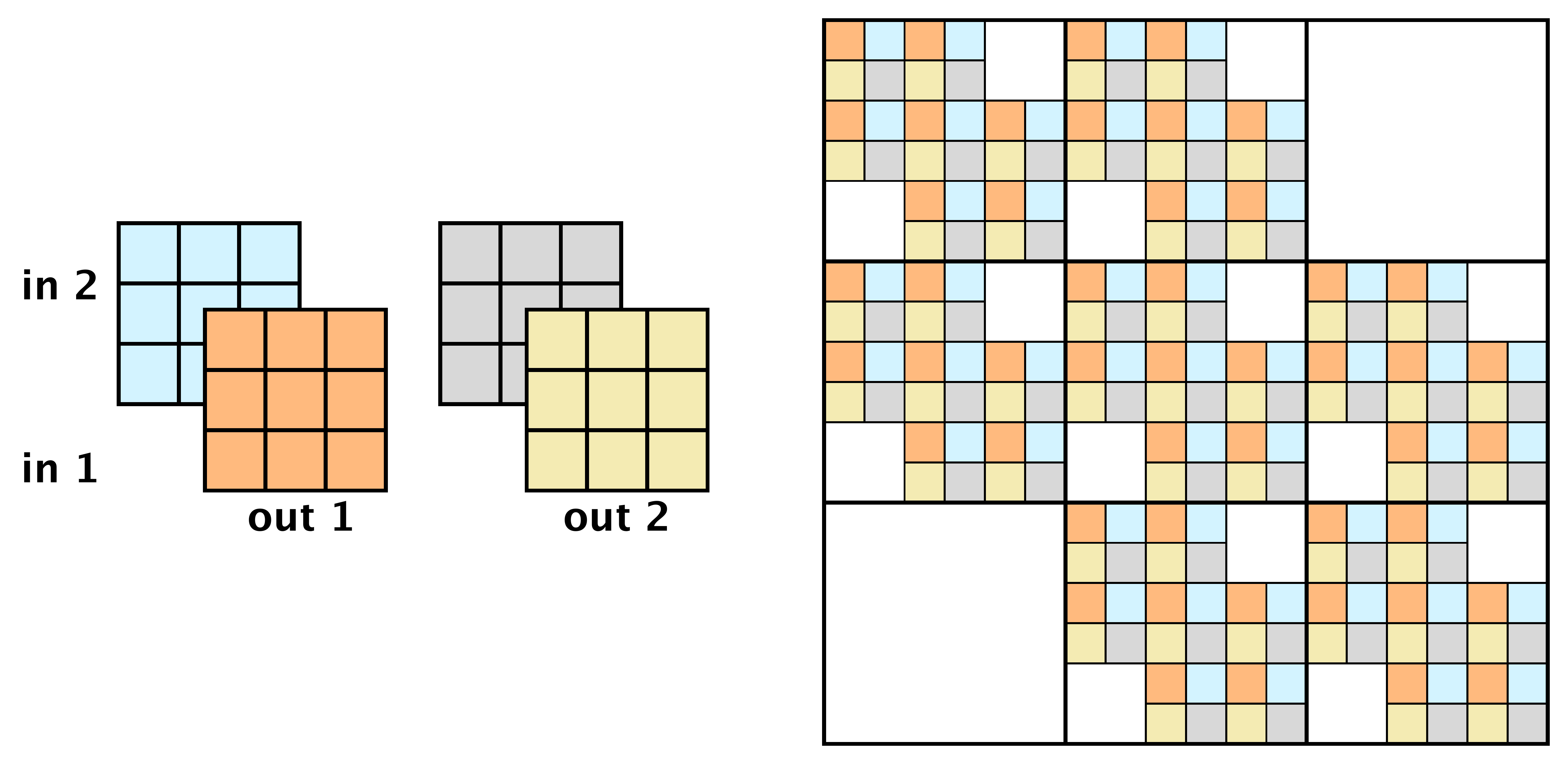}
    \caption{A standard 3 $\times$ 3 convolution layer with two input and output channels. The input is $3 \times 3$ spatially, and has two channels. The convolution uses one-pixel-wide zero padding at each border. Left: the convolution filter $\wvec$. Right: the matrix $\wmat$ which produces the equivalent result when multiplied with a vectorized input.}
    \label{fig:convolution_equivalent_matrix}
\end{figure}

Generally, a convolution layer\footnote{\label{note:conv}In deep learning, convolutions are often actually cross-correlations. In equations, $\star$ denotes a cross-correlation and $*$ denotes a convolution. Moreover, a convolution layer is usually implemented as an aggregation of cross-correlations, i.e. a cross-correlation layer, which is denoted as $\star_l$. In text we may omit these details.} with filter $\wvec$ and input $\xvec$ is equivalent to the multiplication of $\wmat$, a $h \, w \, n_{c_{out}} \times h \, w \, n_{c_{in}}$ matrix, and a vectorized input $\vec{\xvec}$. An example of a single channel convolution and its equivalent matrix is depicted in Figure \ref{fig:convolution_equivalent_matrix_1channel}. The signals $\vec{\xvec}$ and $\vec{\zvec}$ are indexed as $t = i + w \cdot j$, where $i$ is the width index, $j$ is the height index, and $w$ is the total width. Note that the matrix $\wmat$ becomes sparser as the image dimensions grow and that the parameters of the filter $\wvec$ occur repeatedly in the matrix $\wmat$. A two-channel convolution is visualized in Figure \ref{fig:convolution_equivalent_matrix}, where we have omitted parameters inside filters to avoid clutter. Here, $\vec{\xvec}$ and $\vec{\zvec}$ are vectorized using indexing $t = c + n_c \cdot i + (n_c \cdot w) \cdot j$, where $c$ denotes the channel index and $n_c$ the number of channels.

Using standard convolutions as a generative flow is inefficient. The determinant and inverse can be obtained na\"{i}vely by operating directly on the corresponding matrix, but this would be very expensive, corresponding to computational complexity $\mathcal{O}(h^3 \cdot w^3 \cdot n_c^3)$.

\subsection{Autoregressive Convolutions}
%!TEX root = ../main.tex

\begin{figure}
    \centering
    \includegraphics[width=0.40\textwidth]{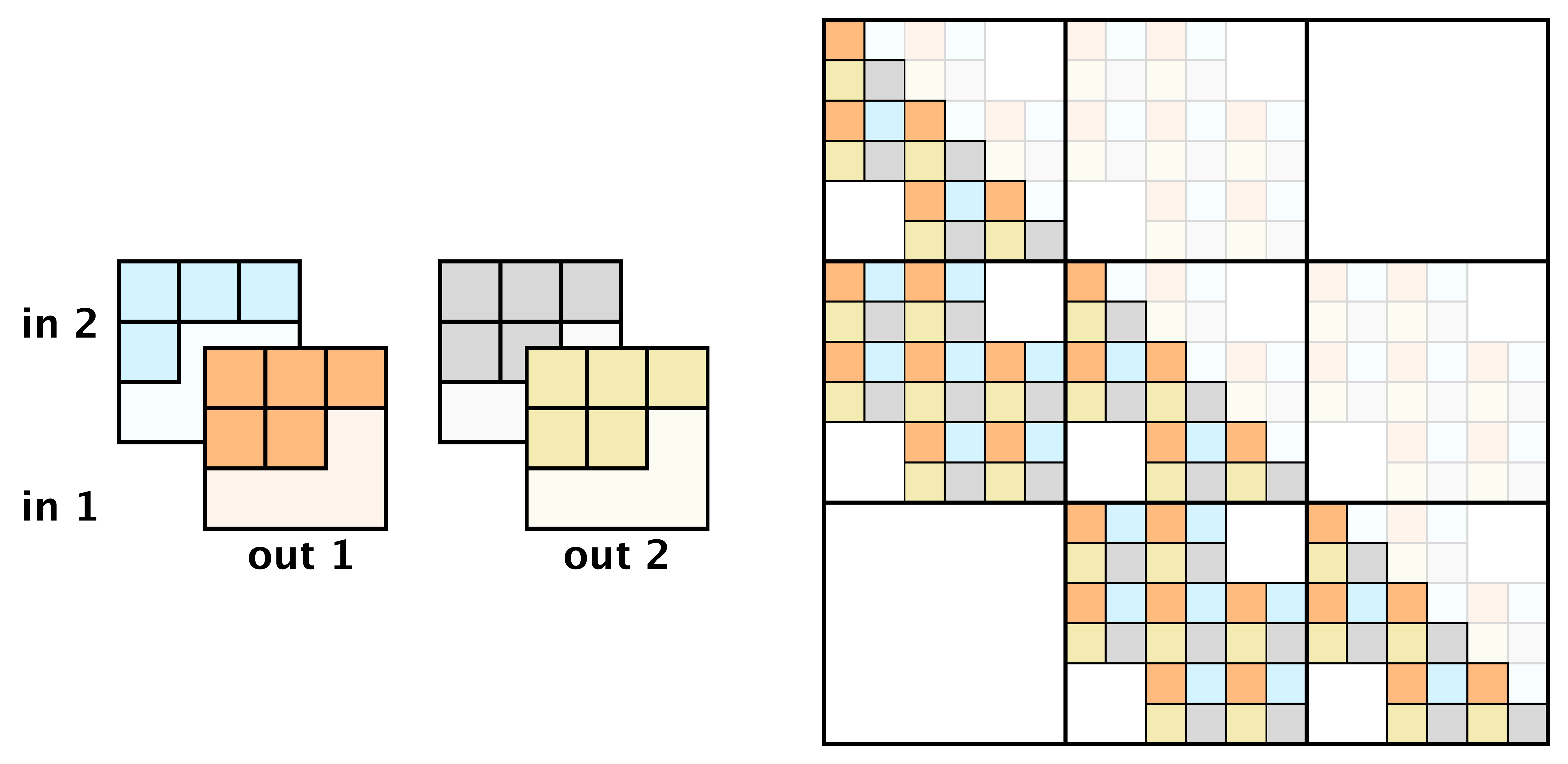}
    \caption{An \textit{autoregressive} 3 $\times$ 3 convolution layer with two input and output channels. The input has spatial dimensions $3 \times 3$, and two channels. The convolution uses one-pixel-wide zero padding at each border. Left: the autoregressive convolution filter $\kvec$. Right: the matrix $\kmat$ which produces the equivalent result on a vectorized input. Note that the equivalent matrix is \textit{triangular}.}
    \label{fig:ar_convolution_equivalent_matrix}
\end{figure}

Autoregressive convolutions have been widely used in the field of normalizing flows \cite{germain2015made, kingma2016improved} because it is straightforward to compute their Jacobian determinant. Although there exist autoregressive convolutions with different input and output dimensions, we let $n_{c_{out}} = n_{c_{in}}$ for invertibility. In this case, autoregressive convolutions can be expressed as a multiplication between a triangular weight matrix and a vectorized input.

In practice, a filter $\kvec = \wvec \odot \mvec$ is constructed from weights $\wvec$ and a binary mask $\mvec$ that enforces the autoregressive structure (see Figure \ref{fig:ar_convolution_equivalent_matrix}). The convolution with the masked filter is autoregressive without the need to mask inputs, which allows parallel computation of the convolution layer:
\begin{equation}
    \zvec = \kvec \star_l \xvec,
    \label{eq:autoregressive_convolution}
\end{equation}
where $\star_l$ denotes a convolution layer\footnoteref{note:conv}. The matrix multiplication $\vec{\zvec} = \kmat \vec{\xvec} $ produces the equivalent result, where $\vec{\xvec}$ and $\vec{\zvec}$ are the vectorized signals, and $\kmat$ is a sparse triangular matrix constructed from $\kvec$ (see Figure \ref{fig:ar_convolution_equivalent_matrix}). The Jacobian is triangular by design and its determinant can be computed in $\mathcal{O}(n_c)$ since it only depends on the diagonal elements of the matrix $\kmat$:
\begin{equation}
    \log \left| \text{det }  \frac{\partial \zvec}{\partial \xvec} \right|
    = h \cdot w \, \sum_c^{n_c} \log \left| k_{c,c,m_y,m_x} \right|,
    \label{eq:autoregressive_logdet}
\end{equation}
where index $c$ denotes the channel and ($m_y$, $m_x$) denotes the spatial center of the filter. The inverse of an autoregressive convolution can theoretically be computed using $\vec{\xvec} = \kmat^{-1} \vec{\zvec}$. In reality this matrix is large and impractical to invert. Since $\kmat$ is triangular, the solution for $\vec{\xvec}$ can be found through forward substitution:
\begin{equation}
    \vec{x}_t = \frac{\vec{z}_t - \sum_{i=1}^{t-1} K_{t,i} \cdot \vec{x}_i }{K_{t,t}} .
    \label{eq:forward_substitution}
\end{equation}
The inverse can be computed by sequentially traversing through the input feature map in the imposed autoregressive order. The computational complexity of the inverse is $\mathcal{O}(h \cdot w \cdot n_c^2)$ and computation can be parallelized across examples in the minibatch.

\section{Method}
\label{sec:method}
%!TEX root = ../main.tex

We present two methods to generalize 1 $\times$ 1 convolutions to invertible $d \times d$ convolutions, improving the flexibility of generative flow models. Emerging convolutions are obtained by chaining autoregressive convolutions (section \ref{sec:emerging_convolutions}), and periodic convolutions are decoupled in frequency domain (section \ref{sec:invertible_periodic_convolutions}). In section \ref{sec:stable_1x1}, we provide a stable and flexible parameterization for invertible $1 \times 1$ convolutions.

\subsection{Emerging convolutions}
\label{sec:emerging_convolutions}
%!TEX root = ../main.tex

Although autoregressive convolutions are invertible, their transformation is restricted by the imposed autoregressive order, enforced through masking of the filters (as depicted in Figure \ref{fig:ar_convolution_equivalent_matrix}). To alleviate this restriction, we propose emerging convolutions, which are more flexible and nevertheless invertible. Emerging convolutions are obtained by chaining specific autoregressive convolutions, invertible via the autoregressive inverses. To some extent this resembles the combination of stacks used to resolve the blind spot problem in conditional image modeling with PixelCNNs \cite{van2016conditional}, with the important difference that we do not constrain the resulting convolution itself to be autoregressive. 

\begin{figure}
    \centering
    \includegraphics[width=0.4\textwidth]{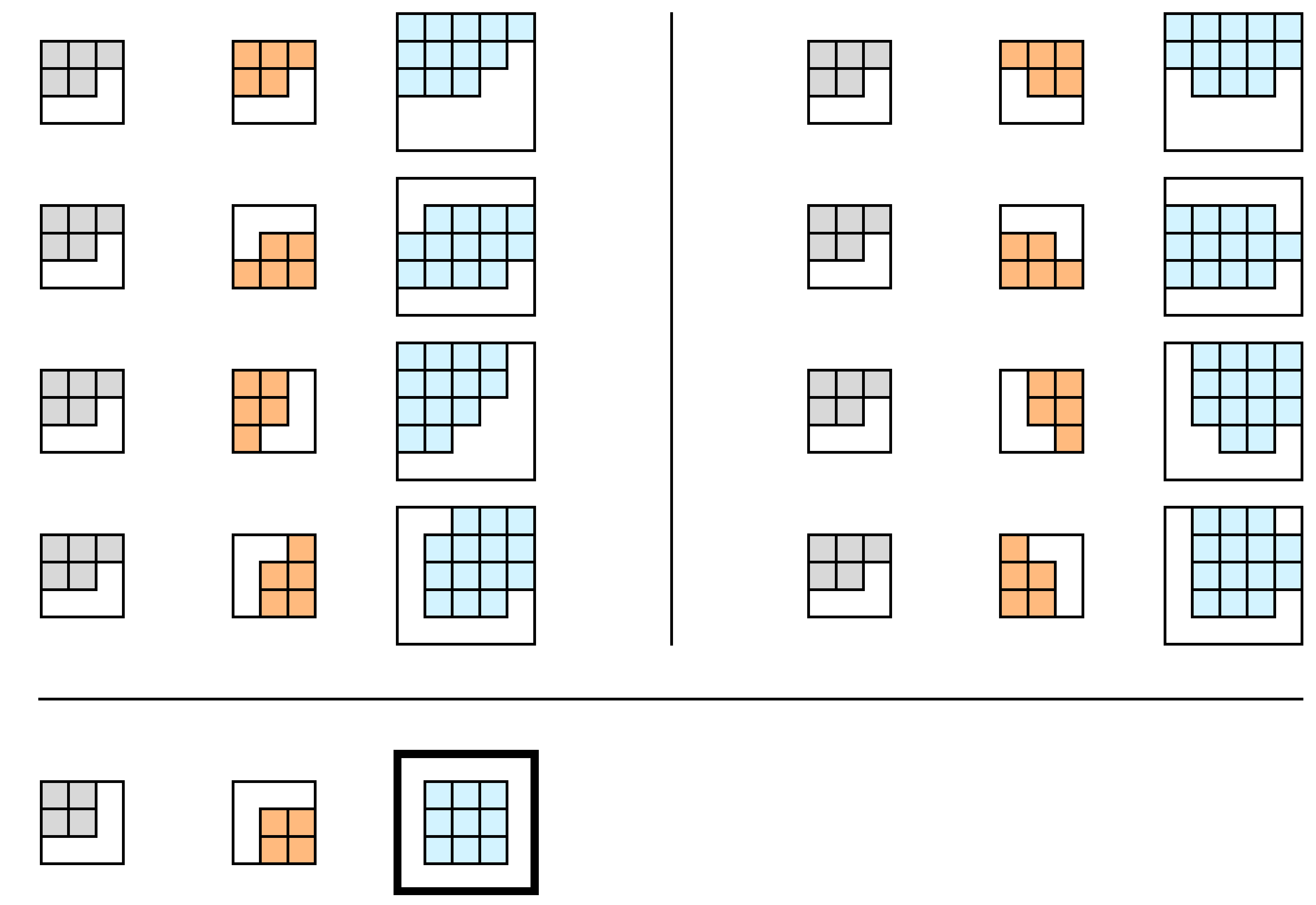}
    \caption{Achievable emerging receptive fields that consist of two distinct auto-regressive convolutions. Grey areas denote the first convolution filter and orange areas denote the second convolution filter. Blue areas denote the emerging receptive field, and white areas are masked. The convolution in the bottom row is a special case, which has a receptive field identical to a standard convolution.}
    \label{fig:emerging_convolutions}
\end{figure}

The emerging receptive field can be controlled by chaining autoregressive convolutions with variations in the imposed order. A collection of achievable receptive fields for emerging convolutions is depicted in Figure \ref{fig:emerging_convolutions}, based on commonly used autoregressive masking.  

The autoregressive inverse requires the solution to a sequential problem, and as a result, it inevitably suffers some additional computational cost. In emerging convolutions we minimize this cost through the use of an accelerated parallel inversion module, implemented in Cython, and by maintaining relatively small dimensionality in the emerging convolutions compared to the internal size of coupling layers. 

\subsubsection{Square Emerging Convolutions}

Deep learning applications tend to use square filters, and libraries are specifically optimized for these shapes. Since most of the receptive fields in Figure \ref{fig:emerging_convolutions} are unusually shaped, these would require masking to fit them in rectangular arrays, leading to unnecessary computation. 

However, there is a special case in which the emerging receptive field of two specific autoregressive convolutions is identical to a standard convolution. These \textit{square} emerging convolutions can be obtained by combining off center square convolutions, depicted in the bottom row of Figure \ref{fig:emerging_convolutions} (also Figure \ref{fig:overview}). Our square emerging convolution filters are more efficient since they require fewer masked values in rectangular arrays.

There are two approaches to efficiently compute square emerging convolutions during optimization and density estimation: \textit{i)} a $d \times d$ emerging convolution is expressed as two smaller consecutive $\frac{d+1}{2} \times \frac{d+1}{2}$ convolutions. Alternatively, \textit{ii)} the order of convolution can be changed: first the smaller $\frac{d+1}{2}$ filters ($k_2$ and $k_1$) are convolved to obtain a single equivalent convolution filter. Then, the output of the emerging convolution is obtained by convolving the equivalent filter, $k = k_2 * k_1$, with the feature map $f$:
\begin{equation}
   k_2 \star (k_1 \star f) = (k_2 * {k_1}) \star f.
   \label{eq:equivalent_emerging_filter}
\end{equation} 
This equivalence follows from the associativity of convolutions and the time reversal of real discrete signals in cross-correlations.

When $d = 1$, two autoregressive convolutions simplify to an LU decomposed $1 \times 1$ convolution. To ensure that emerging convolutions are flexible, we use emerging convolutions that consists of: a single $1 \times 1$ convolution, and two square autoregressive convolutions with different masking as depicted in the bottom row of Figure \ref{fig:overview}. Again, the individual convolutions may all be combined into a single emerging convolution filter using the associativity of convolutions (Equation \ref{eq:equivalent_emerging_filter}).

\subsection{Invertible Periodic Convolutions}
\label{sec:invertible_periodic_convolutions}
%!TEX root = ../main.tex
In some cases, data may be periodic or boundaries may contain roughly the same values. In these cases it may be advantageous to use invertible \textit{periodic} convolutions, which assume that boundaries wrap around. When computed in the frequency domain, this alternative convolution has a tractable determinant Jacobian and inverse. The method leverages the convolution theorem, which states that the Fourier transform of a convolution is given by the element-wise product of the Fourier transformed signals. Specifically, the input and filter are transformed using the Discrete Fourier Transform (DFT) and multiplied element-wise, after which the inverse DFT is taken. By considering the transformation in the frequency domain, the computational complexity of the determinant Jacobian and the inverse are considerably reduced. In contrast with emerging convolutions, which are very specifically parameterized, the filters of periodic convolutions are completely unconstrained.

\noindent A standard convolution layer in deep learning is conventionally implemented as an aggregation of \textit{cross-correlations} for every output channel. The convolution layer with input $\xvec$ and filter $\wvec$ outputs the feature map $\zvec = \wvec \star_l \xvec$, which is computed as:
\begin{equation}
    \zvec_{c_{out}} = \sum_{c_{in}} \wvec_{c_{out}, c_{in}} \star \xvec_{c_{in}}.
\end{equation}
Let $\mathcal{F}(\cdot)$ denote the Fourier transform and let $\mathcal{F}^{-1}(\cdot)$ denote the inverse Fourier transform. The Fourier transform can be moved inside the channel summation, since it is distributive over addition. Let $\hat{\zvec}_{c_{out}} = \mathcal{F}(\zvec_{c_{out}})$, $\hat{\wvec}_{c_{out}, c_{in}} = \mathcal{F}(\wvec_{c_{out}, c_{in}}^*)$ and $\hat{\xvec}_{c_{in}} = \mathcal{F}(\xvec_{c_{in}})$, which are indexed by frequencies $u$ and $v$. Because a convolution differs from a cross-correlation by a time reversal for real signals, let $\wvec_{c_{out}, c_{in}}^*$ denote the reflection of filter $\wvec_{c_{out}, c_{in}}$ in both spatial directions. Using these definitions, each cross-correlation is written as an element-wise multiplication in the frequency domain:
\begin{equation}
    \hat{\zvec}_{c_{out}} = \sum_{c_{in}} \hat{\wvec}_{c_{out}, c_{in}} \odot \hat{\xvec}_{c_{in}},
\end{equation}
which can be written as a sum of products in scalar form:
\begin{equation}
    \hat{z}_{c_{out}, uv} = \sum_{c_{in}} \hat{w}_{c_{out}, c_{in}, uv} \cdot \hat{x}_{c_{in}, uv}.
\end{equation}
The summation of multiplications can be reformulated as a matrix multiplication over the channel axis by viewing the output $\hat{\zvec}_{:,uv}$ at frequency $u,v$ as a multiplication of the matrix $\hat{\wmat}_{uv} = \hat{\wvec}_{:,:, u, v}$ and the input vector $\hat{\xvec}_{:,uv}$:
\begin{equation}
    \hat{\zvec}_{:,uv} = \hat{\wmat}_{uv} \hat{\xvec}_{:,uv}.
    \label{eq:fourier_matrix_multiplication}
\end{equation}
\begin{figure}
    \centering
    \includegraphics[width=0.4\textwidth]{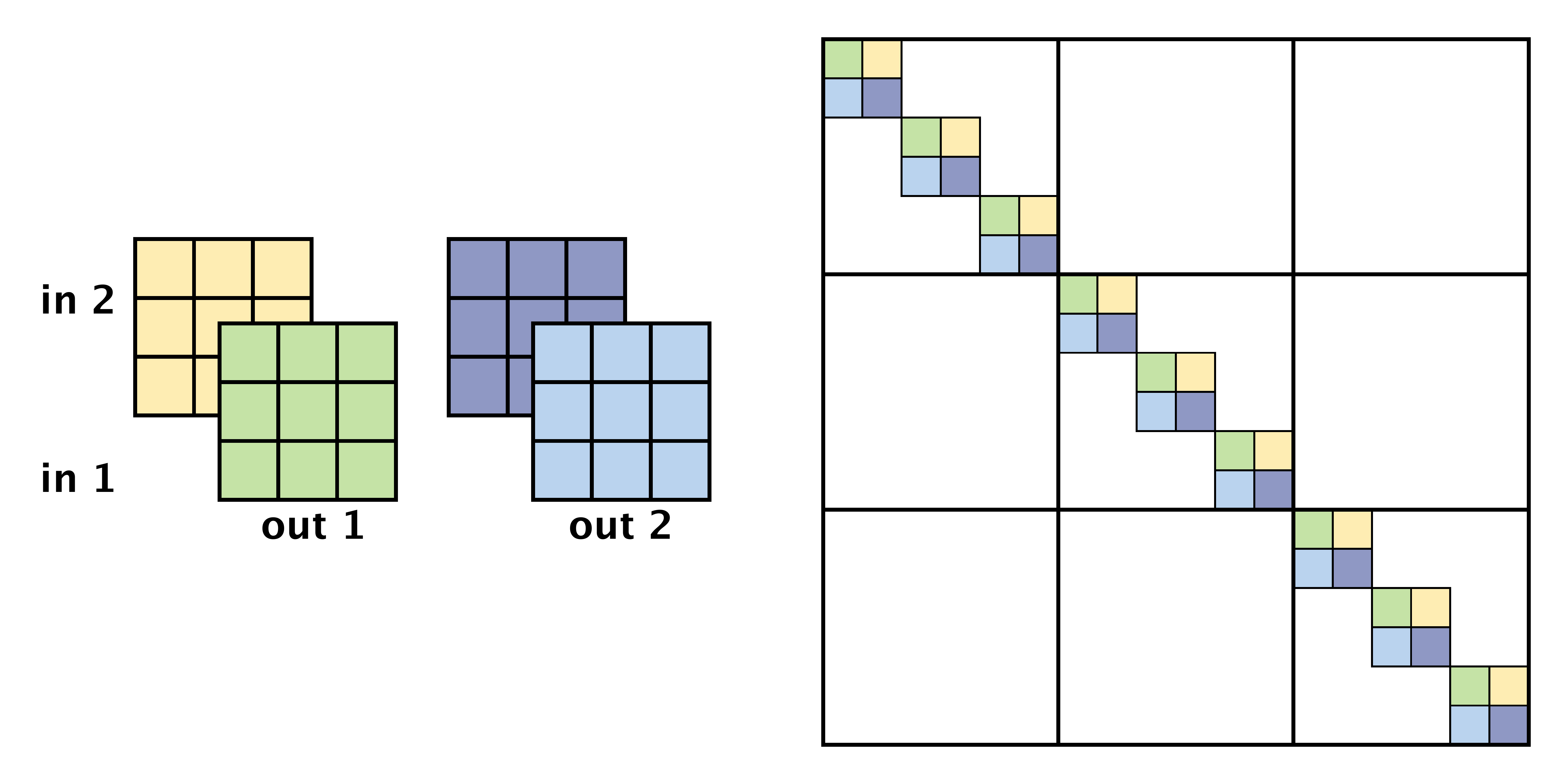}
    \caption{Visualization of a \textit{periodic} 3 $\times$ 3 convolution layer in the \textit{frequency domain}. The input and output have height 3, width 3 and channels
    2. The shape of the filter in the frequency domain determined by the shape of the image, which is also 3 $\times$ 3 spatially in this specific example. Left: the convolution filter transformed to the frequency domain $\hat{\wvec}$. Right: the matrix $\hat{\wmat}$ in the frequency domain, which produces the equivalent result on a vectorized input. The equivalent matrix in the frequency domain is \textit{partitioned}.}
    \label{fig:fourier_convolution_equivalent_matrix}
\end{figure}
The matrix $\hat{\wmat}_{uv}$ has dimensions $c_{out} \times c_{in}$, the input $\hat{\xvec}_{:,uv}$ and output $\hat{\zvec}_{:,uv}$ are vectors with dimension $c_{in}$ and $c_{out}$. The output in the original domain $\zvec_{c_{out}}$ can simply be retrieved by taking the inverse Fourier transform, $\mathcal{F}^{-1}(\hat{\zvec}_{c_{out}})$. The perspective of matrix multiplication in the frequency domain decouples the convolution transformation (see Figure \ref{fig:fourier_convolution_equivalent_matrix}). Therefore, the log determinant of a periodic convolution layer is equal to the sum of determinants of individual frequency components:
\begin{equation}
    \log \left| \text{det } \frac{\partial \zvec}{\partial \xvec} \right| = 
    \log \left| \text{det } \frac{\partial \hat{\zvec}}{\partial \hat{\xvec}} \right| = 
    \sum_{u,v} \log \left| \text{det }  \hat{\wmat}_{uv} \right|.
    \label{eq:fourier_log_det}
\end{equation}
The determinant remains unchanged by the Fourier transform and its inverse, since these are unitary transformations. The inverse operation requires an inversion of the matrix $\hat{\wmat}_{uv}$ for every frequency $u, v$:
\begin{equation}
    \hat{\xvec}_{:,uv} = \hat{\wmat}_{uv}^{-1} \hat{\zvec}_{:,uv}.
    \label{eq:fourier_inverse_frequency}
\end{equation}
The solution of $\xvec$ in the original domain is obtained by the inverse Fourier transform, $\xvec_{c_{in}} = \mathcal{F}^{-1}(\hat{\xvec}_{c_{in}})$, for every channel $c_{in}$. 

In theory, a periodic convolutions may be not invertible, if the determinant of any $\hat{\wmat}_{uv}$ is equal to zero. In practice the filter is initialized with a nonzero determinant. Furthermore, the absolute determinant is maximized in the likelihood objective (Equation \ref{eq:change_of_variables}), which pushes the determinant away from zero.

Recall that a standard convolution layer is equivalent to a matrix multiplication with a $h \, w \, n_{c_{out}} \times h \, w \, n_{c_{in}}$ matrix, where we let $n_{c_{out}} = n_{c_{in}}$ for invertibility. The Fourier transform decouples the transformation of the convolution layer at each frequency, which divides the computation into $h \cdot w$ separate matrix multiplications with $n_c \times n_c$ matrices. Therefore, the computational cost of the determinant is reduced from $\mathcal{O}(h^3 \cdot w^3 \cdot n_c^3)$ to $\mathcal{O}(h \cdot w \cdot n_c^3)$ in the frequency domain, and computation can be parallelized since the matrices are independent across frequencies and independent of the data. Furthermore, the inverse matrices $\hat{\wmat}_{uv}^{-1}$ only need to be computed once after the model has converged, which reduces the inverse convolution to an efficient matrix multiplication with computational complexity\footnote{The inverse also incurs some overhead due to the Fourier transform of the feature maps which corresponds to a computational complexity $\mathcal{O}(h \cdot w \cdot n_c \cdot \log h \, w)$.} $\mathcal{O}(h \cdot w \cdot n_c^2)$.

\subsection{QR 1 $\times$ 1 convolutions}
\label{sec:stable_1x1}
%!TEX root = ../main.tex

Standard 1 $\times$ 1 convolutions are flexible but may be numerically unstable during optimization, causing crashes in the training procedure. \citet{kingma2018glow} propose to learn a PLU decomposition, but since the permutation matrix $\pmat$ is fixed during optimization, their flexibility is limited.

In order to resolve the stability issues while retaining the flexibility of the transformation, we propose to use a \textit{QR} decomposition. Any real square matrix can be decomposed into a multiplication of an orthogonal and a triangular matrix. In a similar fashion to the PLU parametrization, we stabilize the decomposition by choosing $\wmat = \qmat (\rmat + \text{diag}(\svec))$, where $\qmat$ is orthogonal, $\rmat$ is strictly triangular, and elements in $\svec$ are nonzero. 
Any $n \times n$ orthogonal matrix $\qmat$ can be constructed from at most $n$ Householder reflections through  $\qmat = \qmat_1 \qmat_2 \ldots \qmat_{n}$, where $\qmat_i$ is a Householder reflection: 
\begin{equation}
    \qmat_i = \eye - 2 \, \frac{\vvec_i \vvec_i^T }{\vvec_i^T \vvec_i}.
    \label{eq:householder_reflection}
\end{equation}
$\{\vvec_i\}_{i=1}^{n}$ are learnable parameters. Note that in our case $n=n_c$.
In practice, arbitrary flexibility of $\qmat$ may be redundant, and we can trade off computational complexity and flexibility by using a smaller number of Householder reflections. The log determinant of the QR decomposition is $h \cdot w \cdot \text{sum}(\log |\svec|)$ and can be computed in $\mathcal{O}(n_c)$. The computational complexity to construct $\qmat$ is between $\mathcal{O}(n_c^2)$ and $\mathcal{O}(n_c^3)$ depending on the desired flexibility. The QR parametrization has two main advantages: in contrast with the straightforward parameterization it is numerically stable, and it can be completely flexible in contrast with the PLU parametrization.

\section{Related Work}
\label{sec:related_work}
%!TEX root = ../main.tex

The field of generative modeling has been approached from several directions. This work mainly builds upon generative flow methods developed in \cite{rippel2013high, dinh2014nice, dinh2016density, papamakarios2017masked, kingma2018glow}. In \cite{papamakarios2017masked} autoregressive convolutions are also used for density estimation, but both its depth and number of channels makes drawing samples computationally expensive. 

%Our method is compared to these previous works in section \ref{sec:results} using negative log-likelihood (bits/dim). We do not use inception based metrics, as they do not generalize to different datasets, and they do not report overfitting \citep{barratt2018note}. 

Normalizing flows have also been used to perform flexible inference in variational auto-encoders \cite{rezende2015norm, kingma2016improved, tomczak2016improving, van2018sylvester, huang2018neural} and Bayesian neural networks \cite{louizos2017multiplicative}. Instead of designing discrete sequences of transformations, continuous-time normalizing flows can also be designed by drawing a connection with ordinary differential equations \cite{chen2018neural, grathwohl2018ffjord}.

Other likelihood-based methods such as PixelCNNs \citep{van2016pixel} impose a specific order on the dimensions of the image, which may not reflect the actual generative process. Furthermore, drawing samples tends to be computationally expensive. Alternatively, VAEs \citep{kingma2014stochastic} optimize a lower bound of the likelihood. The likelihood can be evaluated via an importance sampling scheme, but the quality of the estimate depends on the number of samples and the quality of the proposal distribution.

Many non likelihood-based methods that can generate high resolution image samples utilize Generative Adversarial Networks (GAN) \citep{goodfellow2014generative}. Although GANs tend to generate high quality images, they do not directly optimize a likelihood. This makes it difficult to obtain likelihoods and to measure their coverage of the dataset.

\section{Results}
\label{sec:results}
%!TEX root = ../main.tex

\begin{figure}[]
    \centering
    \includegraphics[height=3.15cm]{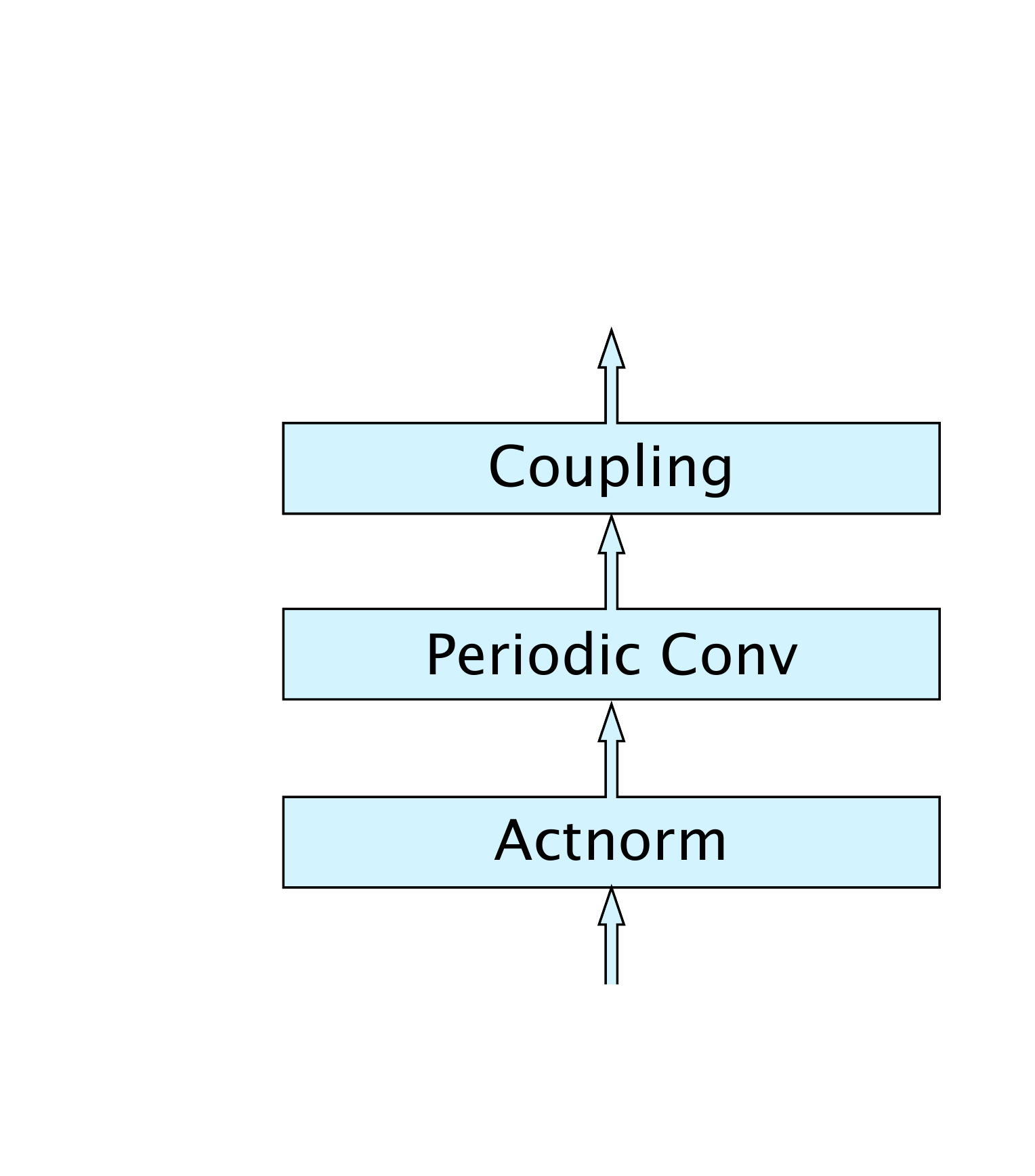}
    \includegraphics[height=3.15cm]{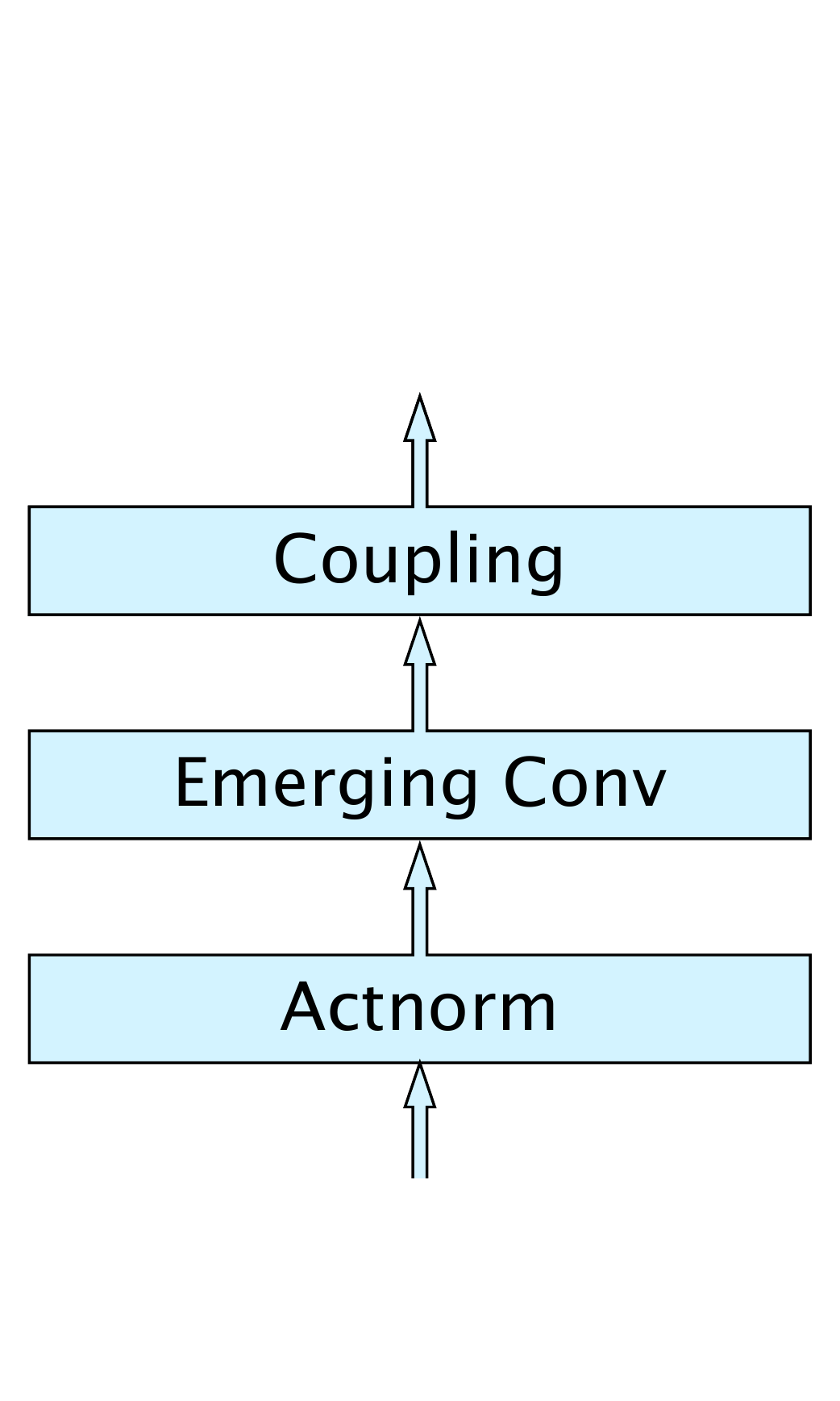}
    \includegraphics[height=3.15cm]{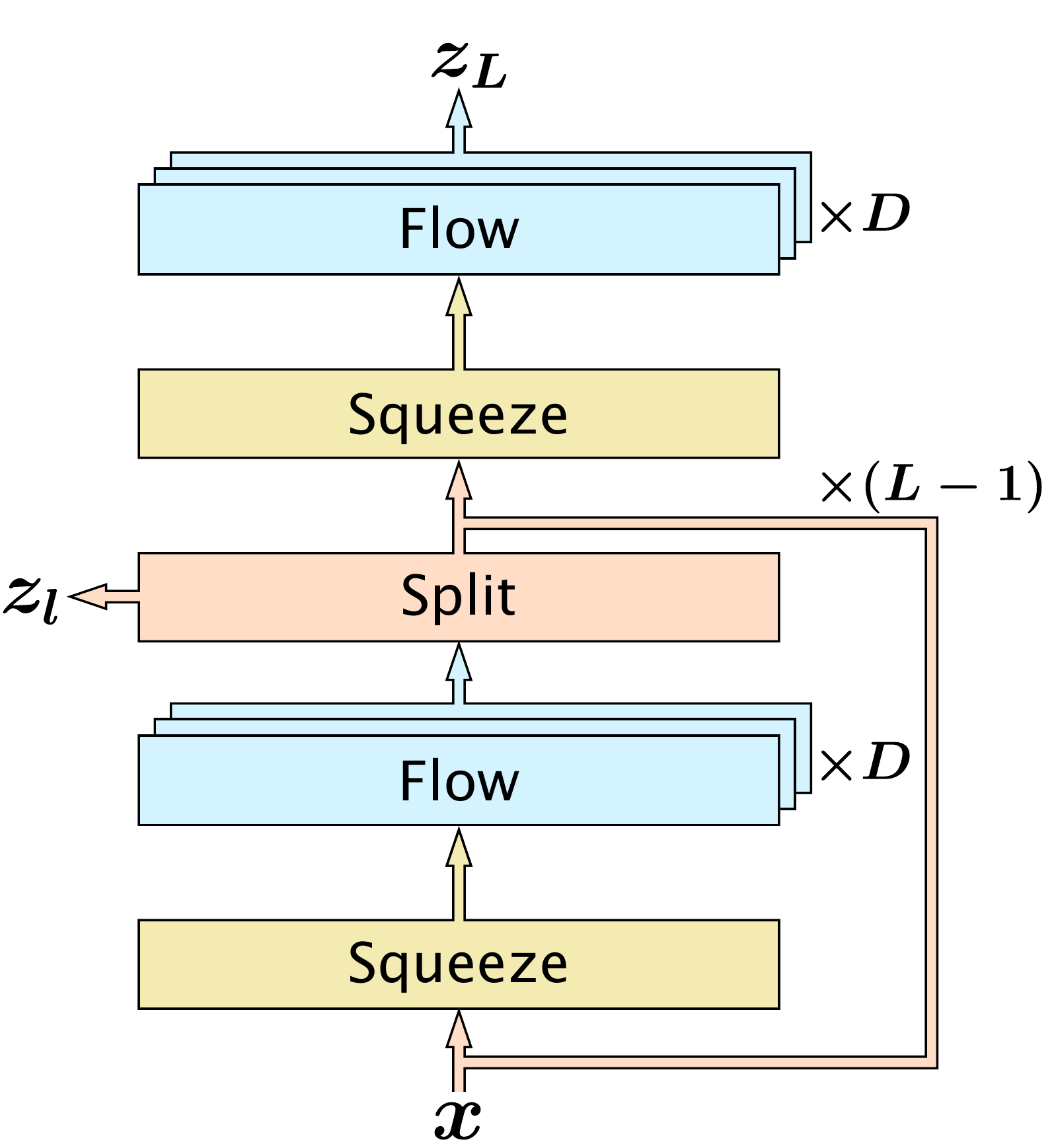}
    \vspace{-0.2cm}
    \caption{Overview of the model architecture. Left and center depict the flow modules we propose: containing either a periodic convolution or an emerging convolution. The diagram on the right shows the entire model architecture, where the flow module is now grouped. The squeeze module reorders pixels by reducing the spatial dimensions by a half, and increasing the channel depth by four. A hierarchical prior is placed on part of the intermediate representation using the split module as in \citep{kingma2018glow}. $x$ and $z$ denote input and output. The model has $L$ levels, and $D$ flow modules per level.}
    \label{fig:model_layout}
\vspace{-0.4cm}
\end{figure}

The architecture of \cite{kingma2018glow} is the starting point for the architecture in our experiments. In the flow module, the invertible $1 \times 1$ convolution can simply be replaced with a $d \times d$ periodic or emerging convolution. For a detailed overview of the architecture see Figure \ref{fig:model_layout}. We quantitatively evaluate models on a variety of datasets in bits per dimension, which is equivalent to the negative log$_2$-likelihood. We do not use inception based metrics, as they do not generalize to different datasets, and they do not report overfitting \citep{barratt2018note}. In addition, we provide image samples generated with periodic convolutions trained on galaxy images, and samples generated with emerging convolutions trained on CIFAR10. 

% Note that generative models are very computationally expensive in general, and we do not have the computational budget to run extremely high-dimensional image modeling tasks.

\subsection{Galaxy density modeling}
\begin{figure}[t]
    \centering
    \includegraphics[width=0.33\textwidth]{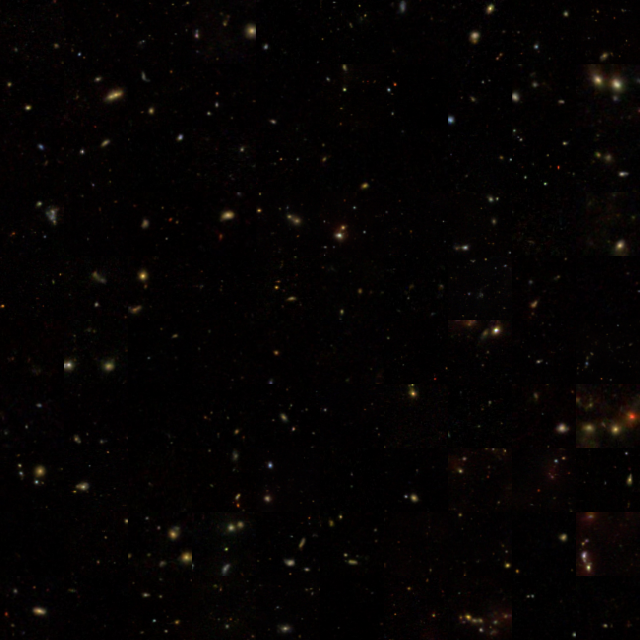}
    \caption{100 samples from a generative flow model utilizing periodic convolutions, trained on the galaxy images dataset.}
    \label{fig:samples_space}
    \vspace{-0.5cm}
\end{figure}

Since periodic convolutions assume that image boundaries are connected, they are suited for data where pixels along the boundaries are roughly the same, or are actually connected. An example of such data is pictures taken in space, as they tend to contain some scattered light sources, and boundaries are mostly dark. \citet{ackermann2018using} collected a small classification dataset of galaxies with images of merging and non-merging galaxies. On the non-merging galaxy images, we compare the bits per dimension of three models, constrained by the same parameter budget: $1 \times 1$ convolutions (Glow), $3 \times 3$ Periodic and $3 \times 3$ Emerging convolutions (see Table \ref{tab:results_space}). Experiments show that both our periodic and emerging convolutions significantly outperform $1 \times 1$ convolutions, and their performance is less sensitive to initialization. Samples of the model using periodic convolutions are depicted in Figure \ref{fig:samples_space}.

\begin{table}
\centering
\caption{Comparison of $1 \times 1$, periodic and emerging convolutions on the galaxy images dataset. Performance is measured in bits per dimension. Results are obtained by running 3 times with different random seeds, $\pm$ reports standard deviation.}
\label{tab:results_space}
\begin{tabularx}{.3\textwidth}{X X }
\hline
& Galaxy  \\ \hline 
$1 \times 1$ (Glow) & 2.03 $\pm 0.026$ \\
Periodic $3 \times 3$ & \textbf{1.98} $\pm 0.003$ \\
Emerging $3 \times 3$ & \textbf{1.98} $\pm 0.007$ \\ \hline 
\end{tabularx}
\end{table}
\begin{figure}[t]
    \centering
    \includegraphics[width=0.33\textwidth]{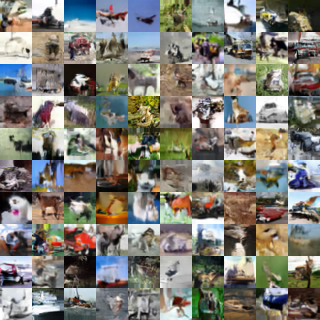}
    \caption{100 samples from a generative flow model utilizing emerging convolutions, trained on CIFAR10.}
    \label{fig:samples_cifar}
    \vspace{-0.5cm}
\end{figure}

\subsection{Emerging convolutions}
The performance of emerging convolution is extensively tested on CIFAR10 \cite{krizhevsky2009learning} and ImageNet \cite{russakovsky2015imagenet}, with different architectural sizes. The experiments in Table \ref{tab:results_emerging} use the architecture from \citet{kingma2018glow}, where emerging convolutions replace the $1 \times 1$ convolutions. Emerging convolutions perform either on par or better than Glow\footnote{The CIFAR10 performance of Glow was obtained by running the code from the original github repository.}, which may be caused by the overparameterization of these large models. Samples of the model using emerging convolutions are depicted in Figure \ref{fig:samples_cifar}. 

In some cases, it may not be feasible to run very large models in production because of the large computational cost. Therefore, it is interesting to study the behavior of models when they are constrained in size. We compare $1 \times 1$ and emerging convolutions with the same number of flows per level ($D$), for $D=8$ and $D=4$. Both on CIFAR10 and ImageNet, we observe that models using emerging convolutions perform significantly better. Furthermore, for smaller models the contribution of emerging convolutions becomes more important, as evidenced by the increasing performance gap (see Table \ref{tab:results_emerging_small}).
\begin{table}
\centering
\caption{Performance of Emerging convolutions on CIFAR10, ImageNet 32x32 and ImageNet 64x64 in bits per dimension (negative log$_2$-likelihood), and $\pm$ reports standard deviation.}
\label{tab:results_emerging}
\begin{tabularx}{0.48\textwidth}{X p{1.8cm} X X }
\hline
& CIFAR10 & ImageNet 32x32 & ImageNet 64x64  \\ \hline 
Real NVP & 3.51 & 4.28 & 3.98 \\
Glow & 3.36 $\pm 0.002$ & \textbf{4.09} & \textbf{3.81} \\
Emerging & \textbf{3.34} $\pm 0.002$ & \textbf{4.09} & \textbf{3.81} \\ \hline 
\end{tabularx}
%\vspace{-0.5cm}
\end{table}
\begin{table}[]
\centering
%\vspace{-0.5cm}
\caption{Performance of Emerging convolutions with different architectures on CIFAR10 and ImageNet 32x32 in bits per dimension. Results are obtained by running 3 times with different random seeds, $\pm$ reports standard deviation.}
\label{tab:results_emerging_small}
\begin{tabularx}{0.47\textwidth}{p{1.7cm} X p{2.3cm} p{.5cm} }
\hline
& CIFAR10 & ImageNet 32x32 & D \\ \hline
$1 \times 1$ (Glow) &  3.46 $\pm 0.005$ & 4.18 $\pm 0.003$ & 8 \\
Emerging & \textbf{3.43} $\pm 0.004$ & \textbf{4.16} $\pm 0.004$ & 8 \\ \hline 
$1 \times 1$ (Glow) & 3.56 $\pm 0.008$ & 4.28 $\pm 0.008$ & 4 \\
Emerging & \textbf{3.51} $\pm 0.001$ & \textbf{4.25} $\pm 0.002$ & 4 \\ \hline 
\end{tabularx}
\end{table}

\subsection{Modeling and sample time comparison with MAF}
Recall that the inverse of autoregressive convolutions requires solving a sequential problem, which we have accelerated with an inversion module that uses Cython and parallelism across the minibatch. Considering CIFAR-10 and the same architecture as used in Table \ref{tab:results_emerging}, it takes 39ms to sample an image using our accelerated emerging inverses, 46 times faster than the na\"{i}vely obtained inverses using tensorflow bijectors (see Table \ref{tab:results_maf}). As expected, sampling from models using $1 \times 1$ convolutions remains faster and takes 5ms.

Masked Autoregressive Flows (MAFs) are a very flexible method for density estimation, and they improve performance over emerging convolutions slightly, 3.33 versus 3.34 bits per dimension. However, the width and depth of MAFs makes them a poor choice for sampling, because it considerably increases the time to compute their inverse: 3000ms per sample using a na\"{i}ve solution, and 650ms per sample using our inversion module. Since emerging convolutions operate on lower dimensions of the data, they are 17 times faster to invert than the MAFs. 

\begin{table}
\centering
\caption{Comparison of $1 \times 1$, MAF and Emerging convolutions on CIFAR-10. Performance is measured in bits per dimension, and the time required to sample a datapoint, when computed in minibatches of size 100. The na\"{i}ve implementation uses Tensorflow bijectors, and our accelerated implementation uses Cython with MPI parallelization.}
\label{tab:results_maf}
\begin{tabularx}{0.48\textwidth}{l p{1.4cm} p{1.7cm} p{1.7cm}}
\hline
CIFAR10 & bits/dim & Na\"{i}ve \newline sample (ms) & Accelerated \newline sample (ms) \\ \hline
$1 \times 1$ (Glow) & 3.36 & \multicolumn{1}{r}{5} & \multicolumn{1}{r}{5} \\
MAF \& $1 \times 1$ & \textbf{3.33} & \multicolumn{1}{r}{3000} & \multicolumn{1}{r}{650} \\
Emerging & 3.34 & \multicolumn{1}{r}{1800} & \multicolumn{1}{r}{39} \\ \hline 
\end{tabularx}
\vspace{-0.1cm}
\end{table}

\subsection{QR 1 $\times$ 1 convolutions}
QR $1 \times 1$ convolutions are compared with standard and PLU convolutions on the CIFAR10 dataset. The models have 3 levels and 8 flows per level. Experiments confirm that our stable QR decomposition achieves the same performance as the standard parameterization, as shown in Table \ref{tab:results_1x1}. This is expected, since any real square matrix has a QR decomposition. Furthermore, the experiments confirm that the less flexible PLU parameterization leads to worse performance, which is caused by the fixed permutation matrix.

\begin{table}[h]
\centering
\vspace{-0.3cm}
\caption{Comparison of standard, PLU and QR $1 \times 1$ convolutions. Performance is measured in bits per dimension (negative log$_2$-likelihood). Results are obtained by running 3 times with different random seeds, $\pm$ reports standard deviation.}
\label{tab:results_1x1}
\begin{tabularx}{0.29\textwidth}{X X }
\hline
Parametrization & CIFAR10 \\ \hline 
W & \textbf{3.46} $\pm 0.005$ \\
PLU & 3.47 $\pm 0.006$ \\
QR & \textbf{3.46} $\pm 0.004$ \\ \hline 
\end{tabularx}
\end{table}

\section{Conclusion}
\label{sec:conclusion}
%!TEX root = ../main.tex

We have introduced three generative flows: \textit{i)} $d \times d$ emerging convolutions as invertible standard zero-padded convolutions, \textit{ii)} $d \times d$ periodic convolutions for periodic data or data with minimal boundary variation, and \textit{iii)} stable and flexible $1 \times 1$ convolutions using a QR parametrization. Our methods show consistent improvements over various datasets using the same parameter budget, especially when considering models constrained in size.

\bibliographystyle{icml2019}
\bibliography{references.bib} 

\begin{thebibliography}{29}
\providecommand{\natexlab}[1]{#1}
\providecommand{\url}[1]{\texttt{#1}}
\expandafter\ifx\csname urlstyle\endcsname\relax
  \providecommand{\doi}[1]{doi: #1}\else
  \providecommand{\doi}{doi: \begingroup \urlstyle{rm}\Url}\fi

\bibitem[Ackermann et~al.()Ackermann, Schawinksi, Zhang, Weigel, and
  Turp]{ackermann2018using}
Ackermann, S., Schawinksi, K., Zhang, C., Weigel, A.~K., and Turp, M.~D.
\newblock Using transfer learning to detect galaxy mergers.
\newblock \emph{Monthly Notices of the Royal Astronomical Society}.

\bibitem[Barratt \& Sharma(2018)Barratt and Sharma]{barratt2018note}
Barratt, S. and Sharma, R.
\newblock A note on the inception score.
\newblock \emph{ICML Workshop on Theoretical Foundations and Applications of
  Deep Generative Models}, 2018.

\bibitem[Chen et~al.(2018)Chen, Rubanova, Bettencourt, and
  Duvenaud]{chen2018neural}
Chen, T.~Q., Rubanova, Y., Bettencourt, J., and Duvenaud, D.~K.
\newblock Neural ordinary differential equations.
\newblock In \emph{Advances in Neural Information Processing Systems}, pp.\
  6572--6583, 2018.

\bibitem[Deco \& Brauer(1995)Deco and Brauer]{deco1995decorr}
Deco, G. and Brauer, W.
\newblock {Higher Order Statistical Decorrelation without Information Loss}.
\newblock In Tesauro, G., Touretzky, D.~S., and Leen, T.~K. (eds.),
  \emph{Advances in Neural Information Processing Systems 7}, pp.\  247--254.
  MIT Press, 1995.

\bibitem[Dinh et~al.(2014)Dinh, Krueger, and Bengio]{dinh2014nice}
Dinh, L., Krueger, D., and Bengio, Y.
\newblock {NICE: Non-linear independent components estimation}.
\newblock \emph{arXiv preprint arXiv:1410.8516}, 2014.

\bibitem[Dinh et~al.(2017)Dinh, Sohl-Dickstein, and Bengio]{dinh2016density}
Dinh, L., Sohl-Dickstein, J., and Bengio, S.
\newblock {Density estimation using Real NVP}.
\newblock \emph{International Conference on Learning Representations, ICLR},
  2017.

\bibitem[Germain et~al.(2015)Germain, Gregor, Murray, and
  Larochelle]{germain2015made}
Germain, M., Gregor, K., Murray, I., and Larochelle, H.
\newblock Made: Masked autoencoder for distribution estimation.
\newblock In \emph{International Conference on Machine Learning}, pp.\
  881--889, 2015.

\bibitem[Gomez et~al.(2017)Gomez, Ren, Urtasun, and
  Grosse]{gomez2017reversible}
Gomez, A.~N., Ren, M., Urtasun, R., and Grosse, R.~B.
\newblock The reversible residual network: Backpropagation without storing
  activations.
\newblock In \emph{Advances in Neural Information Processing Systems}, pp.\
  2214--2224, 2017.

\bibitem[Goodfellow et~al.(2014)Goodfellow, Pouget-Abadie, Mirza, Xu,
  Warde-Farley, Ozair, Courville, and Bengio]{goodfellow2014generative}
Goodfellow, I., Pouget-Abadie, J., Mirza, M., Xu, B., Warde-Farley, D., Ozair,
  S., Courville, A., and Bengio, Y.
\newblock Generative adversarial nets.
\newblock In \emph{Advances in neural information processing systems}, pp.\
  2672--2680, 2014.

\bibitem[Grathwohl et~al.(2018)Grathwohl, Chen, Betterncourt, Sutskever, and
  Duvenaud]{grathwohl2018ffjord}
Grathwohl, W., Chen, R.~T., Betterncourt, J., Sutskever, I., and Duvenaud, D.
\newblock Ffjord: Free-form continuous dynamics for scalable reversible
  generative models.
\newblock \emph{arXiv preprint arXiv:1810.01367}, 2018.

\bibitem[Huang et~al.(2018)Huang, Krueger, Lacoste, and
  Courville]{huang2018neural}
Huang, C.-W., Krueger, D., Lacoste, A., and Courville, A.
\newblock Neural autoregressive flows.
\newblock \emph{arXiv preprint arXiv:1804.00779}, 2018.

\bibitem[Kingma \& Dhariwal(2018)Kingma and Dhariwal]{kingma2018glow}
Kingma, D.~P. and Dhariwal, P.
\newblock {Glow: Generative flow with invertible 1x1 convolutions}.
\newblock In \emph{Advances in Neural Information Processing Systems}, pp.\
  10236--10245, 2018.

\bibitem[Kingma \& Welling(2014)Kingma and Welling]{kingma2014stochastic}
Kingma, D.~P. and Welling, M.
\newblock {Auto-Encoding Variational Bayes}.
\newblock In \emph{Proceedings of the 2nd International Conference on Learning
  Representations}, 2014.

\bibitem[Kingma et~al.(2016)Kingma, Salimans, Jozefowicz, Chen, Sutskever, and
  Welling]{kingma2016improved}
Kingma, D.~P., Salimans, T., Jozefowicz, R., Chen, X., Sutskever, I., and
  Welling, M.
\newblock {Improved variational inference with inverse autoregressive flow}.
\newblock In \emph{Advances in Neural Information Processing Systems}, pp.\
  4743--4751, 2016.

\bibitem[Krizhevsky \& Hinton(2009)Krizhevsky and
  Hinton]{krizhevsky2009learning}
Krizhevsky, A. and Hinton, G.
\newblock Learning multiple layers of features from tiny images.
\newblock Technical report, Citeseer, 2009.

\bibitem[Li \& Grathwohl(2018)Li and Grathwohl]{litraining}
Li, X. and Grathwohl, W.
\newblock {Training Glow with Constant Memory Cost}.
\newblock \emph{NIPS Workshop on Bayesian Deep Learning}, 2018.

\bibitem[Louizos \& Welling(2017)Louizos and
  Welling]{louizos2017multiplicative}
Louizos, C. and Welling, M.
\newblock Multiplicative normalizing flows for variational bayesian neural
  networks.
\newblock In \emph{Proceedings of the 34th International Conference on Machine
  Learning-Volume 70}, pp.\  2218--2227. JMLR. org, 2017.

\bibitem[Papamakarios et~al.(2017)Papamakarios, Murray, and
  Pavlakou]{papamakarios2017masked}
Papamakarios, G., Murray, I., and Pavlakou, T.
\newblock Masked autoregressive flow for density estimation.
\newblock In \emph{Advances in Neural Information Processing Systems}, pp.\
  2338--2347, 2017.

\bibitem[Rezende \& Mohamed(2015)Rezende and Mohamed]{rezende2015norm}
Rezende, D. and Mohamed, S.
\newblock {Variational Inference with Normalizing Flows}.
\newblock In \emph{Proceedings of the 32nd International Conference on Machine
  Learning}, volume~37 of \emph{Proceedings of Machine Learning Research}, pp.\
   1530--1538. PMLR, 2015.

\bibitem[Rezende et~al.(2014)Rezende, Mohamed, and
  Wierstra]{rezende2015stochbackprop}
Rezende, D.~J., Mohamed, S., and Wierstra, D.
\newblock Stochastic backpropagation and approximate inference in deep
  generative models.
\newblock In \emph{Proceedings of the 31st International Conference on Machine
  Learning}, volume~32 of \emph{Proceedings of Machine Learning Research}, pp.\
   1278--1286. PMLR, 2014.

\bibitem[Rippel \& Adams(2013)Rippel and Adams]{rippel2013high}
Rippel, O. and Adams, R.~P.
\newblock High-dimensional probability estimation with deep density models.
\newblock \emph{arXiv preprint arXiv:1302.5125}, 2013.

\bibitem[Russakovsky et~al.(2015)Russakovsky, Deng, Su, Krause, Satheesh, Ma,
  Huang, Karpathy, Khosla, Bernstein, et~al.]{russakovsky2015imagenet}
Russakovsky, O., Deng, J., Su, H., Krause, J., Satheesh, S., Ma, S., Huang, Z.,
  Karpathy, A., Khosla, A., Bernstein, M., et~al.
\newblock Imagenet large scale visual recognition challenge.
\newblock \emph{International journal of computer vision}, 115\penalty0
  (3):\penalty0 211--252, 2015.

\bibitem[Tabak \& Turner(2013)Tabak and Turner]{tabak2013family}
Tabak, E. and Turner, C.~V.
\newblock A family of nonparametric density estimation algorithms.
\newblock \emph{Communications on Pure and Applied Mathematics}, 66\penalty0
  (2):\penalty0 145--164, 2013.

\bibitem[Tabak et~al.(2010)Tabak, Vanden-Eijnden, et~al.]{tabak2010density}
Tabak, E.~G., Vanden-Eijnden, E., et~al.
\newblock Density estimation by dual ascent of the log-likelihood.
\newblock \emph{Communications in Mathematical Sciences}, 8\penalty0
  (1):\penalty0 217--233, 2010.

\bibitem[Theis et~al.(2016)Theis, van~den Oord, and Bethge]{theis2016note}
Theis, L., van~den Oord, A., and Bethge, M.
\newblock A note on the evaluation of generative models.
\newblock In \emph{International Conference on Learning Representations}, 2016.

\bibitem[Tomczak \& Welling(2016)Tomczak and Welling]{tomczak2016improving}
Tomczak, J.~M. and Welling, M.
\newblock Improving variational auto-encoders using householder flow.
\newblock \emph{arXiv preprint arXiv:1611.09630}, 2016.

\bibitem[van~den Berg et~al.(2018)van~den Berg, Hasenclever, Tomczak, and
  Welling]{van2018sylvester}
van~den Berg, R., Hasenclever, L., Tomczak, J.~M., and Welling, M.
\newblock Sylvester normalizing flows for variational inference.
\newblock \emph{arXiv preprint arXiv:1803.05649}, 2018.

\bibitem[van~den Oord et~al.(2016)van~den Oord, Kalchbrenner, Espeholt,
  Vinyals, Graves, et~al.]{van2016conditional}
van~den Oord, A., Kalchbrenner, N., Espeholt, L., Vinyals, O., Graves, A.,
  et~al.
\newblock {Conditional Image Generation with PixelCNN Decoders}.
\newblock In \emph{Advances in Neural Information Processing Systems}, pp.\
  4790--4798, 2016.

\bibitem[Van~Oord et~al.(2016)Van~Oord, Kalchbrenner, and
  Kavukcuoglu]{van2016pixel}
Van~Oord, A., Kalchbrenner, N., and Kavukcuoglu, K.
\newblock Pixel recurrent neural networks.
\newblock In \emph{International Conference on Machine Learning}, pp.\
  1747--1756, 2016.

\end{thebibliography}

% \appendix

% \newpage
% \section{Experimental details}
% \input{sections/details.tex}

\end{document}